\documentclass[11pt]{article}
\usepackage[utf8]{vietnam}
\usepackage[english]{babel}
\usepackage{acl2015}
\usepackage{times}
\usepackage{url}
\usepackage{latexsym}
\usepackage{graphicx}
\usepackage[round]{natbib}
\usepackage[colorlinks=true,linkcolor=blue]{hyperref}%

\usepackage{amssymb}
\usepackage{amsmath}
\usepackage{float}

\graphicspath{ {./figure/}}
\hypersetup{allcolors=blue}

\title{NLPBK at VLSP-2020 shared task: Compose transformer pretrained models for Reliable Intelligence Identification on Social network}

\author{Thanh Chinh Nguyen \\
  Brains Technology, Inc. \\
  {\tt chinh.nguyen@brains-tech.co.jp} \\\And
  Van Nha Nguyen \\
  Websosanh AI \\
  {\tt nhanv@websosanh.org} \\}

\date{}

\begin{document}
\maketitle
\begin{abstract}
This paper describes our method for tuning a transformer-based pretrained model, to adaptation with 
Reliable Intelligence Identification on Vietnamese SNSs problem. We also proposed a model that combines bert-base pretrained models with some metadata features, such as the number of comments, number of likes, images of SNS documents,... to improved results for VLSP shared task: Reliable Intelligence Identification on Vietnamese SNSs. With appropriate training techniques, our model is able to achieve $0.9392$ ROC-AUC on public test set and the final version settles at top 2 ROC-AUC ($0.9513$) on private test set.\end{abstract}

\section{Introduction}

In recent years, the use of SNSs has become a necessary daily activity. As result, SNSs has become the leading tool for spreading news information. In SNSs, News can spread exponentially, but otherwise, a number of users tend to spread unreliable information for their personal purposes affecting the online society. In fact, SNSs has proved to be a powerful source for fake news dissemination (\cite{10.1145/3132847.3132877}, \cite{10.1145/3137597.3137600}). The need for building a system that can identify if news spreading in SNSs is reliable or unreliable is high. However, fact-checking a SNSs post in Vietnamese is several new and challenging research problems. To help handle this problem, VLSP2020's ReINTEL shared task aim participants to build
systems to automatically identify whether an SNSs post is reliable or not (\cite{le2020reintel}). 

We approach this problem as a text classification problem with several specific features of a SNSs post. Accordingly, based on the dataset provided in VLSP2020 shared shared task: Reliable Intelligence Identification on Vietnamese SNSs, we proposed a method that combines between fine-tuning approach to compose several Vietnamese pertrained transformer based models as the main model for text classification purpose, with processing some  meta-data feature like number of likes, number of comment, number of shares, images of the post, ... . We also describe several experiments in fine-tuning the model. Our best model results in a high ROC-AUC of $0.9513$ on the task’s private test and $0.9392$ on the task's public test.

\section{Dataset}

\begin{figure}[!htb]
\caption{Frequency vs length of post message}
\includegraphics[width=\linewidth]{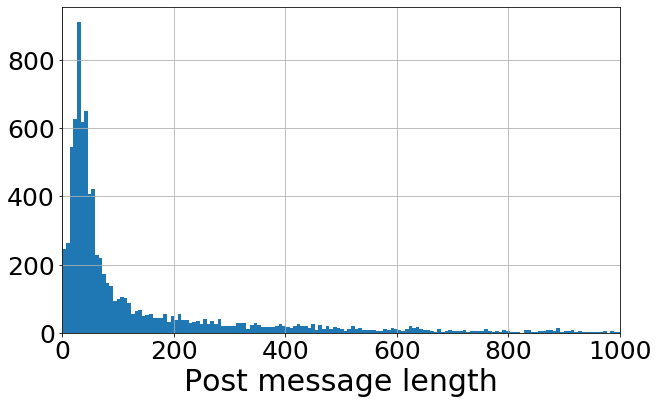}
\label{fig:postLength}
\end{figure}

The dataset was proposed in this task called ReINTEL corpus (\cite{le2020reintel}), consists of a total of 8000 examples split into training, public test, and private test set in 3/1/1 ratio respectively. Each example includes 6 main atributes as follow:
\begin{itemize}
    \item \textbf{user\_name}: the anonymized id of the owner

    \item \textbf{post\_message}: the text content of the news 

    \item \textbf{timestamp\_post}: the time when the news is posted 

    \item \textbf{num\_like\_post}: the number of likes that the news is received

    \item \textbf{num\_comment\_post}: the number of comment that the news is received

    \item \textbf{num\_share\_post}: the number of shares that the news is received
\end{itemize}
Besides, because each example is corresponding to an SNSs post, it may also contain several images that belong to the original port.
A detailed breakdown of data is shown in Table 1. Distribution of post message length in all corpus is illustrated in Figure \ref{fig:postLength}.
\begin{table}[ht]
\label{table1}
\centering
\small
\begin{tabular}{|l|c|c|c|}
\hline
                                    & train &  \begin{tabular}[c]{@{}c@{}}public\\  test\end{tabular}& \begin{tabular}[c]{@{}c@{}}private\\  test\end{tabular} \\ \hline
number of examples                  & 5165  & 1642        & 1646         \\ \hline
average of posts length     & 164   & 148         & 164          \\ \hline
number of posts have images & 1287  & 494         & 508          \\ \hline
number of duplicated posts   & 313   & 31          & 34           \\ \hline
number of duplicated users       & 1464  & 497         & 367          \\ \hline
\end{tabular}
\caption{Detail of ReINTEL Corpus} 
\end{table}

\section{Proposed Model}
The motivation behind our model is the recent success that transfer learning had in a wide range of NLP tasks like text summarization (\cite{liu2019text}, \cite{miller2019leveraging}), relation extraction (\cite{shi2019simple}), text classification (\cite{devlin2019bert}, \cite{sun2020finetune}) , name entity recognition (\cite{devlin2019bert}, \cite{10.1007/978-3-030-51310-8_2}) and  question answering (\cite{devlin2019bert}, \cite{zhang2020retrospective}). The backbone idea of the model is that we adopt a sigmoid classification in front of several transformer-based pretrained models to identify a post message is reliable or not.

\subsection{Transformer based pretrained model}
\begin{figure}[!htb]
\caption{Transformer model architecture}
\includegraphics[width=\linewidth]{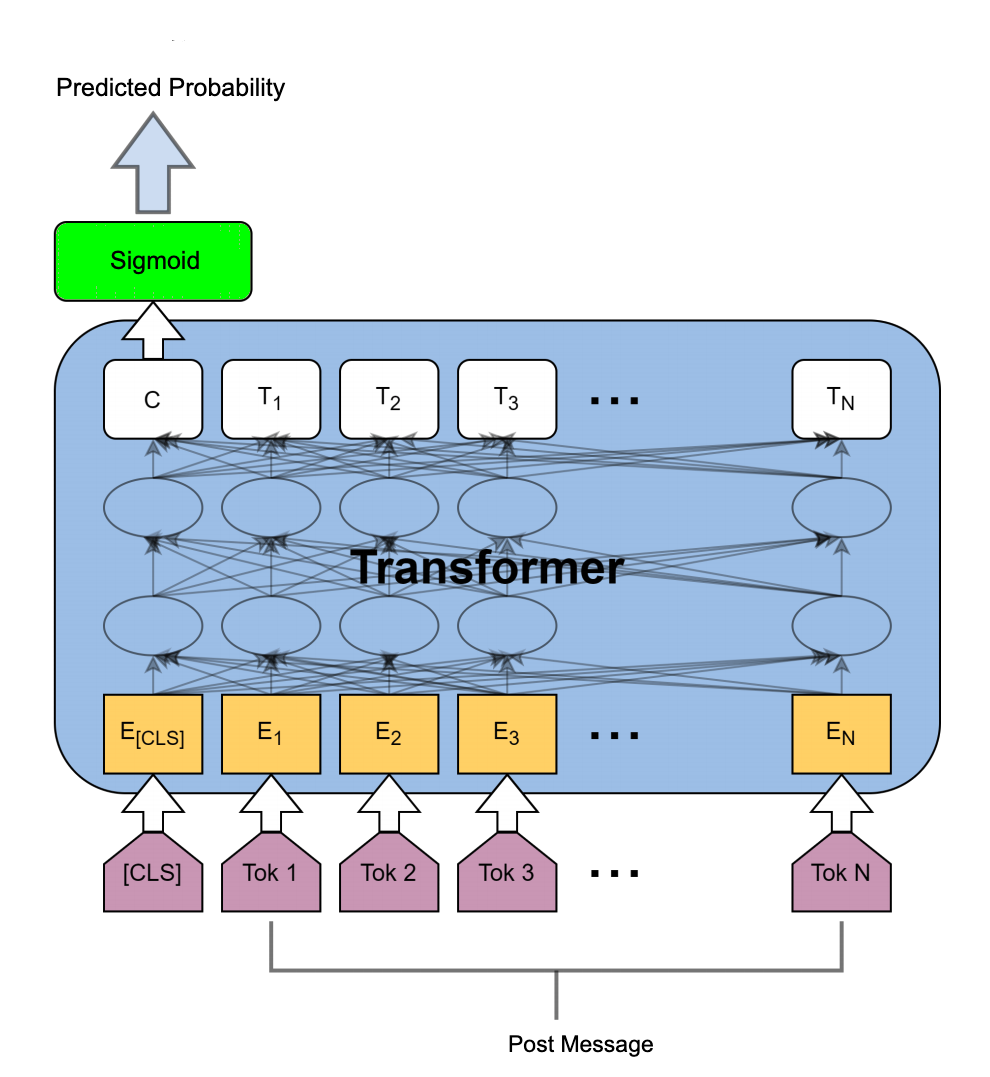}
\label{fig:archi}
\end{figure}

Transformer architectures have been trained on general tasks like language modelling and then can be fine-tuned for another NLP tasks. It takes an input of a sequence and outputs the representations of the sequence. There can be several special segments to flag each important position. For example, in our model, we use \textit{[CLS]} as the first token of the sequence which contains a special classification embedding. The model will take the final hidden state h of this \textit{[CLS]} token as the representation of the whole sequence, a simple sigmoid classification is added to sequence representation in order to predict the label of the whole sequence is reliable or not (\cite{devlin2019bert}). The model architecture sequence classifier is illustrated in Figure \ref{fig:archi}.  

Follow this approach, we used several pretrained transformer models in Vietnamese or in multilingual for this task. 
We selected several models that each model are pretrained in different methods or data domain in order to incorporate more different knowledge from each pre-trained model.(e.g. PhoBERT (\cite{nguyen2020phobert}), XLM (\cite{lample2019crosslingual}), and Bert4News).

Bert4News is our pretrained transformer model for Vietnamese\footnote{We public Bert4News at \url{https://github.com/bino282/bert4news}}. We fine-tuned the original BERT architecture on a $20$ GB tokenized syllables-level Vietnamese news dataset, the model achieves substantial improvements in some Vietnamese NLP tasks. On other hand, PhoBERT (\cite{nguyen2020phobert}) was pretrained on RoBERTa architect (\cite{liu2019roberta}), which can be considered as a variant of BERT (\cite{devlin2019bert}), it removes the Next Sentence Prediction (NSP) task from BERT’s pre-training and introduces dynamic masking so that the masked token changes during the training epochs. PhoBERT is pretrained on a $20$ GB tokenized word-level  Vietnamese corpus. XLM model is a pretrained transformer model for multilingual classification tasks and the use of BERT as initialization of machine translation models on multilingual data crawled from Wikipedia (\cite{lample2019crosslingual}). As result, it can general purpose cross-lingual text representations include Vietnamese text.

\subsection{ Meta data features}

As mentioned above, each example in corpus includes several meta-data of the SNSs post, which will helpful for identify this post is reliable or not. We used several meta data features are provided in the corpus or by preprocessing to improve our model performance:
\begin{itemize}
    \item \textbf{num\_like\_post}: the number of likes that the news is received.

    \item \textbf{num\_comment\_post}: the number of comment that the news is received.

    \item \textbf{num\_share\_post}: the number of shares that the news is received.
    
    \item \textbf{has\_images}: 0 or 1, this post has images or not.
    
    \item \textbf{include\_a\_title}: 0 or 1, we found that sometime, post has include an array of all capital letters (e.g.: \textit{"NẾU LỠ VƯỚNG VIRUSCORONA, BẠN NÊN LÀM GÌ ĐỂ THOÁT HIỂM?..."}), this post will be marked that includes a title.
    
    \item \textbf{day\_in\_year}: \textbf{timestamp\_post} meta data will be format into value that reprensents number of days to this timestamp from 1/1/2020.
\end{itemize}

\subsection{Final model}
In Our model, we generate representations of post message in three methods: tokenized syllables-level text through Bert4News, tokenized word-level text through PhoBERT and tokenized syllables-level text through XLM. We simply concatenate both this three representations with the corresponding post metadata features. This can be considered as a naive model but are proved that can improve performance of systems (\cite{TU2017517}, \cite{supervisedattn}), our results supported this intuition.
\begin{equation}
\small
h^* = h_{Bert4News}\bigoplus h_{PhoBERT}\bigoplus h_{XLM} \bigoplus \xi
\end{equation}
\\
where $h_{Bert4News}$, $h_{PhoBERT}$, $h_{XLM}$, $\xi$ are presentations for Bert4news, PhoBERT and XLM and post metadata features respectively. Final prerentation $h^*$ will be adapted with a sigmoid classifier to get the predicted probability:
\begin{equation}
\small
    \widehat{y} = \sigma(h^*)
\end{equation}
where $\sigma$ represents a sigmoid function. We use  binary cross entropy as our loss function.
\section{Experimental Setup}
ReINTEL corpus includes a lot of examples that have invalid values for several attributes. We filled all these invalid positions with average of valid values of the corresponding attribute, then applied a Min Max Scaler for each attribute in all corpus. With post message that includes a title, we found that title probably be part of sentence, or even a full sentence, so we replace that title with its lower form. Beside, ReINTEL corpus already format web-link to \textit{<URL>} token, and phone number to \textit{<PHONE>} token. We use VnCoreNLP \footnote{\url{https://github.com/vncorenlp/VnCoreNLP}} for tokenizing word-level of message in order to generate PhoBERT representation. Finally, we remove all example that duplicated in both post message, number of likes, number of shares, number of comments and label.

Our model is implemented in pytorch, and use transformer library \footnote{\url{https://github.com/huggingface/transformers}} to load pretrained model. We set batchsize $\beta=16$, learning rate $\alpha=3*10^{-5}$, max sentence length $l=256$. We optimized the model by using k-fold cross-validation, each iteration, 1 fold is used to prevent the models from overfitting while the remaining folds are used for training. Each fold is trained for 5 epochs with early stopping, and save the fold's best model.  We use the average voting strategy to merge the results predicted from all fold' best models. Number of folds are $12$. All hyper parameters are selected based on results of public test. During training, we freeze all layers of pretrained model and just only tuning the adapted sigmoid classification parameters.
\section{Result}
Table 2 reports the results of our models implemented for this shared task on both public test set and private test set. Because of the reason for limiting the number of submissions during the private test phrases, we do not submit the result of the XLM model and the only Bert4news model to the private test set. When fine-tuning with the Bert4News model, we found that adding our metadata can improve the ROC-AUC score up to $1\%$ on the public test set. From the table, our combined model takes the best ROC-AUC score on both public test and private test set among our models, and so, looking at the official rank (\cite{le2020reintel}), we are placed in the middle (2nd from top 3 participants).

\begin{table}[]
\label{result}
\small
\centering
\begin{tabular}{|l|c|c|}
\hline
                     & public test & priavate test \\ \hline
Bert4news  & 0.9271      & x        \\ \hline
Bert4news + metadata & 0.9375      & 0.9503        \\ \hline
PhoBERT + metadata   & 0.9357      & 0.9487        \\ \hline
XLM + metadata       & 0.9298      & x             \\ \hline
proposed model       & \textbf{0.9392}      & \textbf{0.9513}        \\ \hline
\end{tabular}
\caption{Results of our models }

\end{table}

\section{Conclusion}
In this project, we conducted numerous experiments to find a better way of using BERT pretrained weights. We propose a BERT text classification model by combining many different models, aiming to incorporate more knowledge of pretrained models. 

In the future, we would like to further explore the better ways of incorporating task-specific and domain-related knowledge into BERT with in-domain and cross-domain pretraining. Besides, we have not yet taken advantage of information from the images of post in dataset, several previous works are proposed that can help for a fake news detection system (\cite{10.1145/3339252.3341487}), It is also our future work to improve the model performance.

\bibliography{reference}{}

\end{document}